\pdfoutput=1

\documentclass[11pt]{article}

\usepackage{acl}

\usepackage{times}
\usepackage{latexsym}
\usepackage[T1]{fontenc}

\usepackage[utf8]{inputenc}

\usepackage{hyperref}
\usepackage{microtype}
\usepackage{soul}
\usepackage{url}
\usepackage{graphicx}
\usepackage{amsmath}
\usepackage{amsthm}
\usepackage{multirow}
\usepackage{booktabs}
\usepackage{color}
\usepackage{ulem}
\usepackage{algorithm}
\usepackage{algpseudocode}
\usepackage{algorithmicx}
\usepackage[inline]{enumitem}
\usepackage{subfigure}
\usepackage{amssymb}

\newtheorem{theorem}{Theorem}
\newtheorem{lemma}{Lemma}
\newtheorem{definition}{Definition}
\makeatletter
\def\thanks#1{\protected@xdef\@thanks{\@thanks
        \protect\footnotetext{#1}}}
\makeatother

\title{DaMSTF: Domain Adversarial Learning Enhanced Meta Self-Training\\ for Domain Adaptation}

\author{Menglong Lu\textsuperscript{1}$^{\dagger}$\thanks{$\dagger$ contributed equally to this work}, Zhen Huang\textsuperscript{1}$^{\dagger}$, Yunxiang Zhao\textsuperscript{2}$^\ast$, Zhiliang Tian\textsuperscript{1}$^\ast$\thanks{$\ast$ corresponding author}, \\\textbf{Yang Liu}\textsuperscript{1} and \textbf{Dongsheng Li}\textsuperscript{1} \\
        \textsuperscript{1}National Key Laboratory of Parallel and Distributed Computing,\\ National University of Defense Technology, China\\ \textsuperscript{2} Beijing Institute of Biotechnology, China \\
        \small\texttt{\{lumenglong, huangzhen, tianzhiliang, liuyang12a, dsli\}@nudt.edu.cn}, \\
        \small\texttt{zhaoyx1993@163.com}
        }

\begin{document}
\maketitle
\begin{abstract}
Self-training emerges as an important research line on domain adaptation. By taking the model's prediction as the pseudo labels of the unlabeled data, self-training bootstraps the model with pseudo instances in the target domain. 
However, the prediction errors of pseudo labels (label noise) challenge the performance of self-training.
To address this problem, previous approaches only use reliable pseudo instances, i.e., pseudo instances with high prediction confidence, to retrain the model. Although these strategies effectively reduce the label noise, they are prone to miss the hard examples. In this paper, we propose a new self-training framework for domain adaptation, namely Domain adversarial learning enhanced Self-Training Framework (DaMSTF). Firstly, DaMSTF involves meta-learning to estimate the importance of each pseudo instance, so as to simultaneously reduce the label noise and preserve hard examples. Secondly, we design a meta constructor for constructing the meta validation set, which guarantees the effectiveness of the meta-learning module by improving the quality of the meta validation set. Thirdly, we find that the meta-learning module suffers from the training guidance vanishment and tends to converge to an inferior optimal. To this end, we employ domain adversarial learning as a heuristic neural network initialization method, which can help the meta-learning module converge to a better optimal. Theoretically and experimentally, we demonstrate the effectiveness of the proposed DaMSTF. On the cross-domain sentiment classification task, DaMSTF improves the performance of BERT with an average of nearly 4\%.
\end{abstract}

\section{Introduction}
Domain adaptation, which aims to adapt the model trained on the source domain to the target domain, attracts much attention in Natural Language Processing (NLP) applications\cite{du2020adversarial, chen2021wind,lu2022sifter}. Since domain adaptation involves labeled data from the source domain and unlabeled data from the target domain, it can be regarded as a semi-supervised learning problem. From this perspective, self-training, a classical semi-supervised learning approach, emerges a prospective research direction on domain adaptation~\cite{zou2019confidence, liu2021cycle}.  


Self-training consists of a series of loops over the pseudo labeling phase and model retraining phase. In the pseudo labeling phase, self-training takes the model's prediction as the pseudo labels for the unlabeled data from the target domain. Based on these pseudo-labeled instances, self-training retrains the current model in the model retraining phase. The trained model can be adapted to the target domain by repeating these two phases. Due to the prediction errors, there exists label noise in pseudo instances, which challenges self-training approaches~\cite{ZhangST}. 

Previous self-training approaches usually involve a data selection process to reduce the label noise, i.e., preserving the reliable pseudo instances and discarding the remaining ones. In general, higher prediction confidence implies higher prediction correctness, so existing self-training approaches prefer the pseudo instances with high prediction confidence~\cite{zou2019confidence,shin2020two}. However, fitting the model on these easy pseudo instances cannot effectively improve the model, as the model is already confident about its prediction. On the contrary, pseudo instances with low prediction confidence can provide more information for improving the model, but contain more label noise at the same time. 


To simultaneously reduce the label noise and preserve hard examples, we propose to involve in meta-learning to reweight pseudo instances. Within a learning-to-learn schema, the meta-learning module learns to estimate the importance of every pseudo instance, and then, allocates different instance weights to different pseudo instances. Ideally, hard and correct pseudo instances will be assigned larger weights, while easy or error pseudo instances will be assigned smaller weights. To achieve this, the process in the meta-learning module is formulated as a bi-level hyperparameters optimization problem~\cite{franceschi2018bilevel}, where instance weights are taken as the hyperparameters and determined by a series of meta-training steps and meta-validation steps. In the meta-training step, the model is virtually updated on the meta-training set with respect to the current instance weights. In the meta validation step, we validate the virtually updated model with an unbiased meta validation set, and optimize the instance weights with the training guidance back-propagated from the validation performance.

According to the analysis in~\cite{ren2018learning}, a high-quality meta validation set, which is clean and unbiased to the test set, is important for the effectiveness of the meta-learning algorithm. To this end, we propose a meta constructor oriented to the domain adaptation scenario. At each self-training iteration, the meta constructor selects out the most reliable pseudo instances and inserts them into the meta validation set. Since the instances in the meta validation set are all from the target domain and vary along with the self-training iterations,  the data distribution in the constructed meta validation set approximates the one in the target domain. Thus, the meta constructor reduces the bias of the meta validation set. On the other hand, selecting the most reliable pseudo instances can reduce the label noise, making the meta validation set cleaner. 

Another challenge for the meta-learning module is the training guidance vanishment, referring to the gradient vanishment on hyperparameters. With a theoretical analysis, we attribute this problem to the gradient vanishment on the meta validation set. To this end, we introduce a domain adversarial learning module to perturb the model's parameters, thereby increasing the model's gradients on the meta validation set. In DaMSTF, we also interpret the domain adversarial learning module as a heuristic neural network initialization method. Before the model retraining phase, the domain adversarial learning module first initializes the model's parameters by aligning the model's feature space. For domain adaptation, the global optimal refers to the state where the model's parameters are agnostic to the domain information but discriminative to the task information. Thus, the training process in the domain adversarial learning module makes the model's parameters closer to the global optimal, serving as a heuristic neural network initialization.

Our contributions can be summarized as follows:
\begin{itemize}[leftmargin=*]
\item We propose a new self-training framework to realize domain adaptation, named \uline{D}omain \uline{a}dversarial learning enhanced \uline{M}eta \uline{S}elf \uline{T}raining \uline{F}ramework (DaMSTF), which involves meta-learning to simultaneously reduce the label noise and preserve hard examples.
\item We propose a meta constructor to construct the meta validation set, which guarantees the effectiveness of the meta-learning module.

\item  We theoretically point out the training guidance vanishment problem in the meta-learning module and propose to address this problem with a domain adversarial learning module.

\item Theoretically, we analyze the effectiveness of the DaMSTF in achieving domain adaptation. Experimentally, we validate the DaMSTF on two popular models, i.e., BERT for the sentiment analysis task and BiGCN for the rumor detection task, with four benchmark datasets. 
\end{itemize}

\section{Problem Formulation}
\label{subsec:Problem_Setting}

We denote the set that involves all instances in the source domain as $\mathbb{D}_S$, and denote the set that contains all instances in the target domain as $\mathbb{D}_T$. From $\mathbb{D}_S$, we can obtain a labeled dataset for training, i.e., $D_S = \lbrace (x_i, y_i) \rbrace_{i=1}^{N}$. In text classification tasks, the input $x_{i}$ is a text from the input space $\mathcal{X}$, the corresponding label $y_i$ is a $C$-dimensional one-hot label vector, i.e., $y_i  \in \lbrace 0, 1 \rbrace^{C}$, where $C$ is the number of classes.
Based on $D_S$, we learn a hypothesis, $h: \mathcal{X} \to \lbrace 0, 1 \rbrace^{C}$. Since $D_S$ comes from $\mathbb{D}_S$ (i.e., $D_S \subseteq \mathbb{D}_S$), the learned hypothesis $h$ usually performs well on $\mathbb{D}_S$. When we transfer the hypothesis $h$ from $\mathbb{D}_{S}$ to $\mathbb{D}_T$, $h$ may perform poorly due to the domain shift.
The goal of \textit{domain adaptation} is to adapt the hypothesis $h$ to $\mathbb{D}_T$.

In general, unlabeled text in the target domain is available~\cite{gururangan2020don}. We denote the unlabeled target domain dataset as $D_T^{u} = \lbrace (x_m) \rbrace_{m=1}^{U}$, where $x_m \in \mathcal{X}$ is a text input. In some cases, we can even access an \textit{in-domain dataset}, i.e., a small set of labeled data in the target domain, which is denoted as $D_{T}^{l} = \lbrace (x_j, y_j) \rbrace_{j=1}^{L}$ ($x_i \in \mathcal{X}$ and $y_i \in \lbrace 0, 1 \rbrace^{C}$).
When $D_{T}^{l} = \emptyset$, the task is a case of \textit{unsupervised domain adaptation}~\cite{wilson2020a}. Otherwise, the task is a case of \textit{semi-supervised domain adaptation}~\cite{saito2019semi}. 

\section{Methodology}
\label{sec:methodology}
\begin{algorithm}[t]
\small
\caption{DaMSTF}
\label{algo:meta self-training}
\begin{algorithmic}[1]
\Require labeled source dataset $D_S$, unlabeled target dataset $D_T^{u}$, in-domain dataset $D_{T}^{l}$
\State Pretrain $\mathbf{\theta}$ on $D_S$, $D_{M} \leftarrow D_{T}^{l}$
\While{the termination criteria is not met}
\State Compute pseudo label $\hat{\mathbf{Y}}_{T}$ on $D_{T}^{u}$
\State $H = -\hat{\mathbf{Y}}_{T}*log(\hat{\mathbf{Y}}_{T})$ 
\State Sort the $D^p_{T}$ with respect to $H$ in ascending order, and denote the first $\mathcal{K}$ data as $D_E$, the remaining data as $D_T^{tr}$ 
\State $D_{M} = D_{T}^l \cup D_{E}$ 
\State \Call{DomainAdversarial}{$D_S \cup D^{u}_{T}$, $\mathbf{\theta}_F$, $\vartheta$}
\State \Call{MetaLearning}{$D_{S} \cup D^{tr}_{T}$, $\mathbf{\theta}$, $\mathbf{w}$}
\EndWhile
\Function{MetaLearning}{$D$, $\mathbf{\theta}$, $\mathbf{w}$}
\For {training batch $\mathcal{B}$ in $D$}
\For{t=1 $\to \mathcal{T}_{M}$} 
\State Compute $\hat{\mathbf{\theta}}(\mathbf{w}^t)$ via Eq.~\eqref{eq:theta_hat}
\State Compute weight $\mathbf{w}^{t+1}$ via Eq.~\eqref{eq:w_update_A}
\EndFor
\State $\mathbf{w}^{*} \leftarrow \mathbf{w}^{\mathcal{T}_M}$, update $\mathbf{\theta} \textnormal{ with } Eq.~\eqref{eq:theta_meta_update}$
\EndFor
\State \Return{$\mathbf{\theta}$, $\mathbf{w}$}
\EndFunction

\Function{DomainAdversarial}{$D$, $\mathbf{\theta}_{F}$, $\vartheta$}
\For{training batch $\mathcal{B}$ in $D$}
  \For{t=1 $\to \mathcal{T}_{D}$ }
    \State $\vartheta = \vartheta - \eta_{1} \triangledown_{\vartheta}\mathcal{L}_{DA}(\mathbf{\theta}_F, \vartheta, \mathcal{B})$  
  \EndFor
  \For{t=1 $\to \mathcal{T}_{G}$ }
    \State $\mathbf{\theta}_F = \mathbf{\theta}_F + \eta_{2} \triangledown_{\mathbf{\theta}}\mathcal{L}_{DA}(\mathbf{\theta}_F, \vartheta, \mathcal{B})$  
  \EndFor
\EndFor
\State \Return{$\mathbf{\theta}$, $\vartheta$}
\EndFunction
\end{algorithmic}
\end{algorithm}
\subsection{Model Overview}
\label{subsec:overview}
DaMSTF inherits the basic framework of self-training, which consists of iterations over the ``Pseudo Labeling'' phase and the ``Model Retraining'' phase.
To achieve domain adaptation, self-training simultaneously optimizes the model's parameters and the pseudo labels with Eq.~\eqref{eq:self-training}.

\begin{small}
\begin{eqnarray}
\min_{\mathbf{\mathbf{\theta}}, \hat{\mathbf{Y}}_T} \mathcal{L}_{st}(\mathbf{\mathbf{\theta}}, \hat{\mathbf{Y}}_T)\! =\! \sum_{(x_k, y_k) \in D_S}{\mathcal{E}(\Phi(x_k; \mathbf{\theta}), y_k)}\! + \nonumber\\
\!\sum_{x_i \in D_T^u}{\mathcal{E} (\Phi(x_i; \mathbf{\theta}), \hat{y}(x_i))} \label{eq:self-training}
\end{eqnarray}
\end{small}

\noindent where $\hat{\mathbf{Y}}_T = [\hat{y}_1, \hat{y}_2, \ldots, \hat{y}_{|D^u_{T}|}]^{T}$ denotes the pseudo label set of the unlabeled target domain data, $\Phi_{\mathbf{\theta}}$ denotes the model under the hypothesis ($h$), and $\mathbf{\mathbf{\theta}}$ denotes the model's parameters.

In the pseudo labeling phase, DaMSTF predicts the unlabeled data in the target domain, and the predictions are taken as pseudo labels. Then, these pseudo instances are sent to the meta constructor. For the instances with high prediction confidence, the meta constructor uses them to expand the meta validation set. For the remaining ones, the meta constructor uses them to construct the meta-training set.

\begin{figure*}[t]
  \hspace{-10pt}
    \includegraphics[width=5.0in]{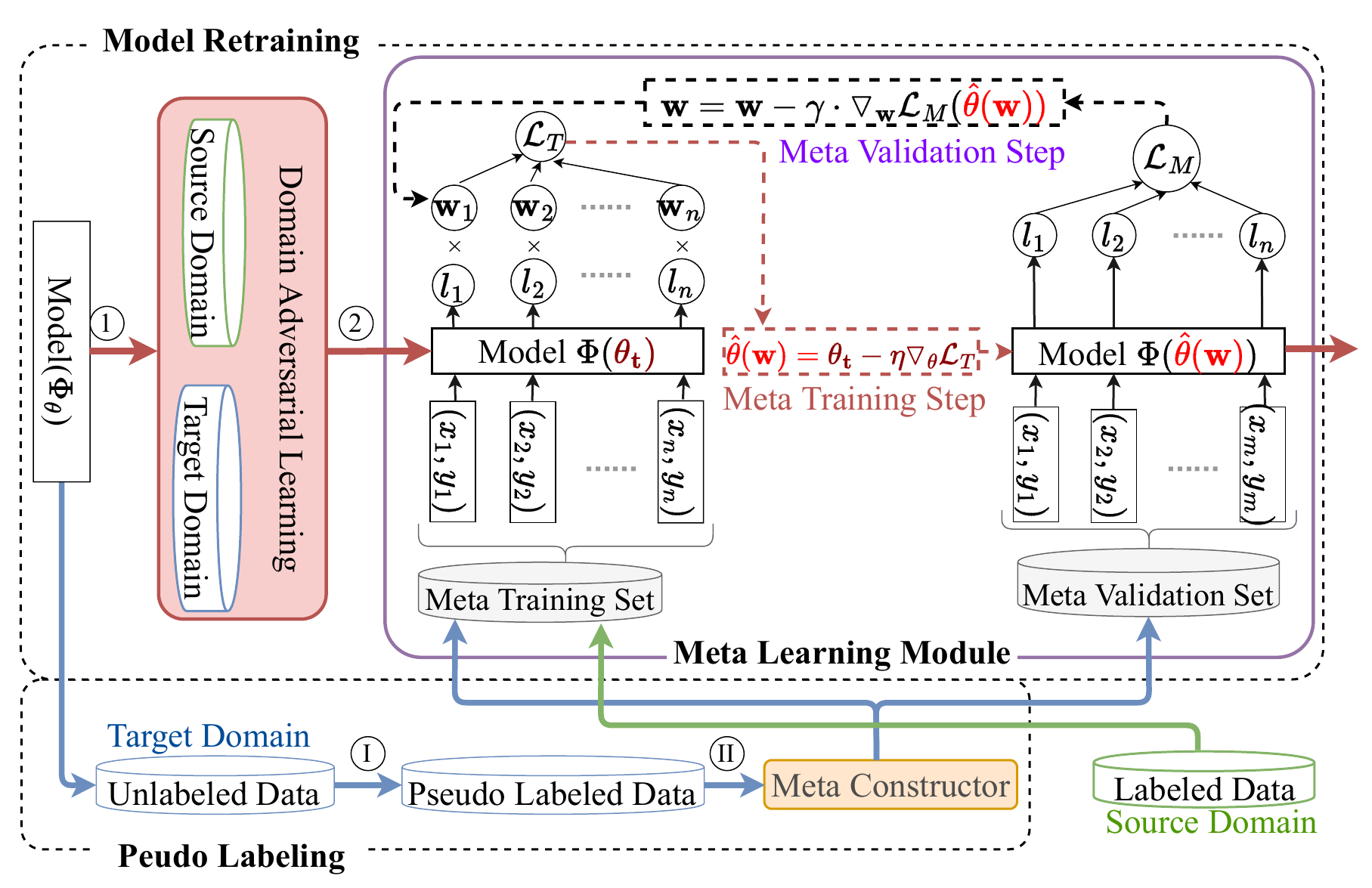}
    \label{subfig:meta self-train}
  \centering
  \caption{An overview of the DaMSTF. Red arrows indicate the training process of the model, while blue and green 
 arrows indicate the data flow.
  }
  \label{fig:overview}
\end{figure*}

In the model retraining phase, DaMSTF first trains the model in the domain adversarial training module to align the feature space. Then, the model is trained in the meta-learning module. Afterward, DaMSTF backs to the pseudo labeling phase to start another self-training iteration. 

Fig.~\ref{fig:overview} shows the structure of DaMSTF, and Algorithm~\ref{algo:meta self-training} presents the corresponding pseudo-code.

\subsection{Meta-Learning Module}
\label{subsec:meta_learning_module}
As described in Fig.~\ref{fig:overview}, the meta-learning module involves a series of loops over the ``Meta Training'' step and ``Meta Validation'' step to optimize the hyper-parameters and the model parameters.

  
  

\noindent\textbf{Meta Training.} The training batch in the meta training phase, i.e., $\mathcal{B} = \lbrace (x_1, y_1), (x_2, y_2), \ldots \rbrace$, merges the labeled data from the source domain with the pseudo labeled data from the target domain. The supervision on the pseudo instances is the pseudo-label, and the supervision on the labeled instances is the ground-truth label. 
We compute the risk loss on the training batch with Eq.~\eqref{eq:inner}:

\begin{small}
\begin{eqnarray}
\mathcal{L}_{T}(\mathbf{\theta}, \mathbf{w}^t, \mathcal{B}) &=& \frac{1}{|\mathcal{B}|}\sum_{x_i, y_i \in \mathcal{B}}{\sigma(\mathbf{w}^t_i) \mathcal{E}(\Phi(x_i; \mathbf{\theta}), y_i)} \label{eq:inner} 
\end{eqnarray}
\end{small}

\noindent where $|\mathcal{B}|$ is the size of $\mathcal{B}$, $\mathcal{E}$ is the loss function. $\Phi_{\mathbf{\theta}}$ denotes the model under the hypothesis ($h$), and $\mathbf{\mathbf{\theta}}$ denotes the model's parameters. $\mathbf{w}_1, \mathbf{w}_{2}, \ldots, \mathbf{w}_{|\mathcal{B}|}$ are the extra hyperparameters introduced in the meta-learning module, i.e., a set of instance weights indicating the importance of each training example. $\sigma$ represents the sigmoid function, which scales the instance weights into $[0, 1]$. In the meta training step, we derive a virtual update on the model with Eq.~\eqref{eq:theta_hat}:

\begin{small}
\begin{eqnarray}
\hat{\mathbf{\theta}}(\mathbf{w}^t) &=& \mathbf{\theta} - \eta \triangledown_{\mathbf{\theta}} \mathcal{L}_{T}(\mathbf{\theta}, \mathbf{w}^t, \mathcal{B}) \label{eq:theta_hat} 
\end{eqnarray}
\end{small}
where $\eta$ is the learning rate.

\noindent \textbf{Meta Validation} After being virtually updated in the meta training phase,  the model is validated on the meta validation set $D_M$ with Eq.~\eqref{eq:L_meta}:

\begin{small}
\begin{eqnarray}
\mathcal{L}_{M}(\hat{\mathbf{\theta}}(\mathbf{w}^t)) = \frac{1}{|D_M|} \cdot \sum_{x_j, y_j \in D_{M}}{\mathcal{E}(\Phi(x_j;\hat{\mathbf{\theta}}(\mathbf{w}^t)), y_j)} \label{eq:L_meta}
\end{eqnarray}
\end{small}

\noindent where $\mathcal{E}$ is the loss function, $|D_M|$ is the size of the meta validation set. By backpropagating the performance on the meta validation set, we derive the \textit{training guidance} for updating the instance weights on the training batch as below:

\begin{small}
\begin{eqnarray}
\frac{\partial \mathcal{L}_M(\hat{\mathbf{\theta}}(\mathbf{w}))}{\partial \mathbf{w}} &=& \frac{\partial{\mathcal{L}_M(\hat{\mathbf{\theta}}(\mathbf{w}))}}{\partial{\hat{\mathbf{\theta}}(\mathbf{w})}}\cdot\frac{\partial \hat{\mathbf{\theta}}(\mathbf{w})}{\partial{\mathbf{w}}} \label{eq:w_guidance}
\end{eqnarray}
\end{small}

To reduce the computation cost, we use the approximation technique in~\cite{chen2021wind} to compute the training guidance (i.e., $\frac{\partial\mathcal{L}_M(\hat{\mathbf{\theta}}(\mathbf{w}))}{\partial \mathbf{w}}$).




Based on the computed training guidance, we obtain the optimal instance weights (marked as $\mathbf{w}^{*}$) with gradient descent algorithm, as described in Eq.~\eqref{eq:w_update_A}. Further, we update $\theta$ with Eq.~\eqref{eq:theta_meta_update}:

\begin{small}
\begin{eqnarray}
\mathbf{w}^{t+1}=& \mathbf{w}^t - \gamma \cdot \frac{\partial \mathcal{L}_M(\hat{\mathbf{\theta}}(\mathbf{w}))}{\partial \mathbf{w}} \label{eq:w_update_A} \\
\mathbf{\theta}^{t+1}=& \mathbf{\theta}^t - \eta \triangledown_{\mathbf{\theta}} \mathcal{L}_{T}(\mathbf{\theta}, \mathbf{w}^*, \mathcal{B}) \label{eq:theta_meta_update}
\end{eqnarray}
\end{small}

After the above process is completed on the training batch $\mathcal{B}$, another training batch will be selected to start the meta-learning phase again, as shown in lines 15-21 in Algorithm~\ref{algo:meta self-training}.

\subsection{Meta Constructor}
\label{subsec:data_selection}
In previous studies, the meta validation set is constructed by collecting a set of labeled data that have the same distribution as the test set~\cite{ren2018learning,shu2019meta}. However, such practice is not acceptable in domain adaptation, as we are not aware of the data distribution of the target domain during the training phase. 

To this end, we propose a meta constructor to construct a meta validation set that approximates the target domain. Specifically, we select the reliable instances from the pseudo-labeled data as the instances in the meta validation set. 
To evaluate the reliability of each of the pseudo instances, we compute their prediction entropy via Eq.~\eqref{eq:entropy}:

\begin{small}
\begin{equation}
\label{eq:entropy}
H(x_i) = -\sum_{c=1}^{C} (\Phi(c|x_i;\mathbf{\theta}) \cdot log(\Phi(c|x_i;\mathbf{\theta})))
\end{equation}
\end{small}

\noindent where $\Phi(c|x_i;\mathbf{\theta})$ is the probability of the instance $x_i$ belongs to the $c_{th}$ category. 

In general, a lower prediction entropy indicates a higher prediction correctness~\cite{nguyen2020ensemble}. Thus, we first sort the $D_T^p$ (pseudo labeled dataset) in ascending order according to their prediction entropy. Then, the top-ranked $\mathcal{K}$ instances, denoted as $D_E$, are selected as the validation instances, and the remaining pseudo samples, denoted as $D_T^{tr}$, are preserved in the meta training set. 

In the semi-supervised domain adaptation, we take the in-domain dataset to initialize the meta validation dataset and use $D_E$ to expand the meta validation set along with the self-training iterations. In the unsupervised domain adaptation, where the in-domain dataset is empty, we directly take $D_E$ as the meta validation set. The above process is detailed in lines 2-8 of Algorithm~\ref{algo:meta self-training}.

Here, meta constructor is an important knot that combines meta-learning and self-training. On the one hand, traditional machine learning approaches cannot exploit the pseudo instances with high prediction entropy, due to the inherent label noise. In this case, the meta constructor uses them to construct the meta training set, as the meta-learning module is tolerant to the label noise in the meta-training set.  On the other hand, pseudo instances with low prediction entropy cannot provide extra information for improving the model but contain less label noise. In this case, the meta constructor uses them to validate the model, i.e., uses them to construct or expand the meta validation set, which can improve the quality of the meta validation set.

\subsection{Domain Adversarial Learning}
\label{subsec:domain_adver_learning}

As theoretically explained in $\S$~\ref{subsec:val_exp}, the training guidance would not be indicative if the model's gradient on the validation instance is negligible. The presence of domain adversarial learning can prevent the gradient vanishment on the meta validation set, thereby preventing the training guidance vanishment. On the other hand, domain adversarial learning can explicitly align the feature space along with the self-training iterations.

To present the details in the domain adversarial learning module, we divide the model $\Phi({\bullet;\mathbf{\theta}})$ into two parts: the feature extraction layer $\Phi_F(\bullet;\mathbf{\theta}_F)$ and the task-specific layer $\Phi_c(\bullet;\mathbf{\theta}_c)$. Usually, $\mathbf{\theta}_c$ is the parameters of the last layer in the model, whose output is the prediction probability of each category. The prediction process in the model is:

\begin{small}
\begin{equation}
\Phi(x_i; \mathbf{\theta}) = \Phi_c(\Phi_F(x_i;{\mathbf{\theta}_F});\mathbf{\theta}_c)
\end{equation}
\end{small}

Following~\citet{ganin2016domain}, we introduce an extra domain discriminator to discriminate the instances' domains, i.e., $\mathbf{\varphi}(\bullet;\mathbf{\vartheta})$, where $\vartheta$ is the parameters. On a training batch $\mathcal{B}$, the risk loss for domain adversarial learning is: 

\begin{small}
\begin{eqnarray}
\mathcal{L}_{DA}(\mathbf{\theta}_F, \vartheta, \mathcal{B}) = \frac{1}{|\mathcal{B}|} \sum_{x_i, d_i \in \mathcal{B}}\mathcal{E}(\varphi(\Phi_F(x_i;\mathbf{\theta}_F);\vartheta), d_i) 
\end{eqnarray}
\end{small}

\noindent where $d_i$ is a one-hot vector representing the domain of $x_i$, $\mathcal{E}$ is the cross-entropy function. The specific training process of the proposed domain adversarial learning module is depicted in Algorithm~\ref{algo:meta self-training}, lines 25-35. 

\section{Theoretical Analysis}
\label{sec:theoretical_analysis}

This section first introduces the training guidance vanishment problem and then explains the effectiveness of DaMSTF in achieving domain adaptation. The proofs are detailed in Appendix.~\ref{sec:appendix_A} and Appendix.~\ref{sec:appendix_B}.

\subsection{Training Guidance Vanishment}
\label{subsec:val_exp}
\begin{theorem}
\label{theo:1}
Let $\mathbf{w}_{i}$ be the weight of the training instance $i$, denoted as $(x_i, y_i)$, in $\mathcal{B}$, the gradient of $\mathbf{w}_{i}$ on $\mathcal{L}_M$ can be represented by the similarity between the gradients on training instance $i$ and the gradients on the meta validation set: 

\begin{small}
\begin{equation}
\frac{\partial L_M(\hat{\mathbf{\theta}}(\mathbf{w}))}{\partial \mathbf{w}_i} = -\frac{\eta}{|\mathcal{B}|}\cdot [\frac{1}{|D_M|}\sum_{j=1}^{|D_M|} \vec{\mathbf{g}}_{\hat{\mathbf{\theta}}}(x_j, y_j)^T] \cdot \vec{\mathbf{g}}_{\mathbf{\theta}}(x_i, y_i) \nonumber
\end{equation}
\end{small}

\noindent where $\frac{1}{|D_M|}\sum_{j=1}^{|D_M|} \vec{\mathbf{g}}_{\hat{\mathbf{\theta}}}(x_j, y_j)^T$ is the gradients of $\hat{\mathbf{\theta}}$ on $D_M$, $\vec{\mathbf{g}}_{\mathbf{\theta}}^i(x_i, y_i)$ is the gradients of $\mathbf{\theta}$ on the training instance $i$, $\eta$ is the learning rate in Eq.~\eqref{eq:theta_hat}
\end{theorem}

According to Theorem~\ref{theo:1}, $\frac{\partial L_M(\hat{\mathbf{\theta}}(\mathbf{w}))}{\partial \mathbf{w}_i}$ is not indicative for every training instance if the model's gradient on the meta validation set (i.e., $\frac{1}{|D_M|}\sum_{j=1}^{|D_M|}\vec{\mathbf{g}}_{\hat{\mathbf{\theta}}}(x_j, y_j)$) is very small, which we named as the \textit{training guidance vanishment} problem. In DaMSTF, the meta-learning module is challenged by the training guidance vanishment problem from the following aspects. 

Firstly, the meta validation set is much smaller than the meta training set, so the model converges faster on the meta validation set than that on the meta training set. Considering the optimization on neural networks is non-convex, the model can converge to an inferior optimal if it converges too early on the meta validation set. In this case, the model's gradient on the meta validation set is very small, which results in the training guidance vanishment. 

Secondly, the instances in $D_{E}$ are the ones with small prediction entropy. Since the supervision for the pseudo instances is exactly the model's predictions, lower prediction entropy results in lower risk loss. Then, the gradients back-propagated from the risk loss are negligible, which also results in the training guidance vanishment. 


\subsection{Theoretical Explanation of DaMSTF}
\label{subsec:4.2}
The \textit{disagreement} and $H{\Delta}H$-distance were first proposed in~\citet{ben2010theory} and have been widely applied to analyze the effectiveness of domain adaptation approaches~\cite{saito2019semi,du2020adversarial}. For any two different hypotheses $h_1$ and $h_2$, \textit{disagreement} $\epsilon_D(h_1, h_2)$ quantifies the discrepancy of their different predictions on a specific dataset $D$. When $h_2$ is an ideal hypothesis that can correctly map all instances in $D$, $\epsilon_D(h_1, h_2)$ also represents the \textit{error rate} of the hypothesis $h_1$ on dataset $D$, abbreviated as $\epsilon_D(h_1)$.
$H{\Delta}H$-distance is a metric for evaluating the divergence of the data distribution between two datasets, which is only relevant to the input space of the datasets. 


\begin{theorem}
\label{theo:2}
Assume there exists an ideal hypothesis, denoted as $h^*$, which correctly maps all instances in the target domain to their groud-truth labels. In the self-training iteration $t$, let $\epsilon_{D_{T}^{l}}(h^t)$ and $\epsilon_{D_E}(h^t)$ be the error rate of the hypothesis $h^{t}$ on $D^l_{T}$ and $D_E$, respectively. Then, the error rate of the hypothesis $h^{t}$ on the target domain is upper bounded by: 

\begin{small}
\begin{eqnarray}
\epsilon_{\mathbb{D}_T}(h^t) \leq \epsilon_{D_{T}^{l} \cup D_E}(h^t) + \frac{1}{2}d_{H \Delta H}(\mathbb{D}_T, D_{T}^{l} \cup D_E) \nonumber \\
  + \rho \cdot \epsilon_{D_E}(h^*, h^{t-1}) \nonumber
\end{eqnarray}
\end{small}

\noindent where $\rho = \frac{|D_E|}{|D_{T}^{l}|+|D_E|}$ is a coefficient related to the size of $D^l_{T}$ and $D_{E}$, $\epsilon_{D_{T}^{l} \cup D_E}(h^t)$ is the error rate of the hypothesis $h^{t}$ on the union of $D_{T}^{l}$ and $D_E$.
\end{theorem}
\begin{theorem}
\label{theo:3}
Assume there exists three datasets, $D_{1}$, $D_{2}$, $D_{3}$,  and let $X_1$, $X_2$, $X_3$ denotes the set of input cases in these three datasets, i.e., $X_1 = \lbrace x_i|(x_i, y_i) \in D_1 \rbrace$, $X_2 = \lbrace x_i|(x_i, y_i) \in D_2 \rbrace$, $X_3 = \lbrace x_i|(x_i, y_i) \in D_3 \rbrace$. If $X_{1}\subseteq X_{2} \subseteq X_{3}$, then 
$$d_{H\Delta H}(D_2, D_3) \leq d_{H\Delta H}(D_1, D_3)$$
holds
\end{theorem}

Based on Theorem~\ref{theo:2}, we demonstrate the effectiveness of DaMSTF from the following aspects. 

First of all, expanding the meta validation set can decrease the second term in Theorem~\ref{theo:2}, i.e., $\frac{1}{2}d_{H \Delta H}(\mathbb{D}_T, D_{T}^{l} \cup D_E)$. According to Theorem~\ref{theo:3}, $d_{H \Delta H}(\mathbb{D}_T, D_{T}^{l} \cup D_E)$ is smaller than $d_{H \Delta H}(\mathbb{D}_T, D_{T}^{l})$, as the input cases in $D_E$ and $D_T^l$ are all belong to the input cases in the $\mathbb{D}_T$. Thus, expanding the meta validation set can reduce the upper bound of $\epsilon_{\mathbb{D}_T}(h^t)$

What's more, as $D_E$ varies in each self-training iteration, the DaMSTF can leverage the diversity of the unlabeled data in the target domain. Thus, $d_{H \Delta H}(\mathbb{D}_T, D_{T}^{l} \cup D_E)$ is close to $d_{H \Delta H}(\mathbb{D}_T, D_T^u)$ in the whole training process.

Last but not least, by selecting examples that have the lowest prediction entropy, the error rate on $D_{E}$ is much lower than that of the expected error rates on $D_{T}^{p}$, formally, $\epsilon_{D_E}(h^*, h^{t-1})<\epsilon_{D_T^p}(h^*, h^{t-1})$. In other words, the data selection process in the meta constructor reduces the third term in Theorem~\ref{theo:2},i.e., $\rho \cdot \epsilon_{D_E}(h^*, h^{t-1})$.  

\section{Experiments}
\label{sec:experiments}
We provide the experiment settings in $\S$~\ref{subsec:exp_settings} and compare DaMSTF with previous domain adaptation approaches in $\S$~\ref{subsec:exp_General_Results}. In $\S$~\ref{subsec:Ablation_Study}, we analyze the effectiveness of the meta constructor and the domain adversarial learning module with an ablation study. $\S$~\ref{subsec:unlabeled data} validate that exposing more unlabeled data to DaMSTF can improve the domain adaptation performance (Theorem~\ref{theo:3}). Appendix~\ref{sec:extra_exp} provides extra experiments of the domain adversarial learning module in preventing the training guidance vanishment problem, and the meta-learning module in highlighting the hard and correct pseudo instances.

\subsection{Experiment Settings}
\label{subsec:exp_settings}
\paragraph{Dataset}
On the rumor detection task, we conduct experiments with the public dataset TWITTER~\cite{zubiaga2016learning}. As the instances in the TWITTER dataset are collected with five topics, we categorized the instances into five domains. 
On the sentiment classification task, we conduct experiments  withs the public dataset Amazon~\cite{blitzer2007biographies}. 
We follow the method in~\cite{he2018adaptive} to preprocess the Amazon dataset, and the resultant dataset consists of 8,000 instances from four domains: books, dvd, electronics, and kitchen. More statistics about the TWITTER dataset and the Amazon dataset can be found in Appendix~\ref{sec:dataset}. 

\paragraph{Implementation Details}The base model on the rumor detection task is BiGCN~\cite{bian2020rumor}, while the base model on the sentiment classification task is BERT~\cite{devlin2019bert}. On the benchmark datasets, we conduct domain adaptation experiments on every domain. When one domain is taken as the target domain for evaluation, the rest domains are merged as the source domain. 
More impelementation details are provided in Appendix~\ref{sec:appendix_C}.

\begin{table*}[t]
\small
\centering
\begin{tabular}{c|cccc|ccccc}
\hline
\multirow{2}{*}{\begin{tabular}[c]{@{}c@{}}Target \\ Domain\end{tabular}} & \multicolumn{4}{c|}{Unsupervised domain adaptation}                  & \multicolumn{5}{c}{Semi-Supervised domain adaptation}                        \\ \cline{2-10} 
                                                                          & Out            & DANN  & \multicolumn{1}{c|}{CRST}  & DaMSTF         & In+Out         & MME   & BiAT  & \multicolumn{1}{c|}{Wind}  & DaMSTF         \\ \hline
Cha.                                                                      & \textit{0.561} & 0.501 & \multicolumn{1}{c|}{0.563} & \textbf{0.635} & \textit{0.586} & 0.601 & 0.547 & \multicolumn{1}{c|}{0.552} & \textbf{0.649} \\ \cline{2-10} 
Fer.                                                                      & \textit{0.190} & 0.387 & \multicolumn{1}{c|}{0.446} & \textbf{0.524} & \textit{0.200} & 0.081 & 0.256 & \multicolumn{1}{c|}{0.291} & \textbf{0.629} \\ \cline{2-10} 
Ott.                                                                      & \textit{0.575} & 0.544 & \multicolumn{1}{c|}{0.709} & \textbf{0.753} & \textit{0.599} & 0.612 & 0.614 & \multicolumn{1}{c|}{0.633} & \textbf{0.843} \\ \cline{2-10} 
Syd.                                                                      & \textit{0.438} & 0.461 & \multicolumn{1}{c|}{0.673} & \textbf{0.717} & \textit{0.424} & 0.677 & 0.661 & \multicolumn{1}{c|}{0.628} & \textbf{0.731} \\ \hline
Mean                                                                      & \textit{0.441} & 0.473 & \multicolumn{1}{c|}{0.598} & \textbf{0.657} & \textit{0.452} & 0.493 & 0.520 & \multicolumn{1}{c|}{0.526} & \textbf{0.714} \\ \hline
\end{tabular}
\caption{F1 score on the TWITTER}
\label{tab:TWITTERResults}
\end{table*}

\begin{table*}[htb]
\small
\centering
\begin{tabular}{l|llll|lllll}
\hline
\multirow{2}{*}{\begin{tabular}[c]{@{}l@{}}Target\\ Domain\end{tabular}} & \multicolumn{4}{c|}{Unsupervised Domain Adaptation}                  & \multicolumn{5}{c}{Semi-Supervised Domain Adaptation}                        \\ \cline{2-10} 
                                                                         & Out            & DANN  & \multicolumn{1}{l|}{CRST}  & DaMSTF         & In+Out         & MME   & BiAT  & \multicolumn{1}{l|}{Wind}  & DaMSTF         \\ \hline
books                                                                    & \textit{0.882} & 0.887 & \multicolumn{1}{l|}{0.878} & \textbf{0.931} & \textit{0.890} & 0.896 & 0.891 & \multicolumn{1}{l|}{0.890} & \textbf{0.947} \\ \hline
dvd                                                                      & \textit{0.831} & 0.864 & \multicolumn{1}{l|}{0.845} & \textbf{0.917} & \textit{0.882} & 0.893 & 0.888 & \multicolumn{1}{l|}{0.904} & \textbf{0.935} \\ \hline
electronics                                                              & \textit{0.871} & 0.914 & \multicolumn{1}{l|}{0.877} & \textbf{0.925} & \textit{0.918} & 0.906 & 0.926 & \multicolumn{1}{l|}{0.917} & \textbf{0.941} \\ \hline
kitchen                                                                  & \textit{0.863} & 0.922 & \multicolumn{1}{l|}{0.868} & \textbf{0.927} & \textit{0.925} & 0.93  & 0.934 & \multicolumn{1}{l|}{0.933} & \textbf{0.947} \\ \hline
Mean                                                                     & \textit{0.862} & 0.897 & \multicolumn{1}{l|}{0.867} & \textbf{0.925} & 0.904          & 0.906 & 0.910 & \multicolumn{1}{l|}{0.911} & \textbf{0.942} \\ \hline
\end{tabular}
\caption{Macro-F1 score on the Amazon dataset}
\label{tab:AmazonResults}
\end{table*}
\paragraph{Comparing Methods} Since the DaMSTF can be customized to both semi-supervised and unsupervised domain adaptation scenarios, the baselines contain both unsupervised and semi-supervised domain adaptation approaches. For the unsupervised domain adaptation, Out~\cite{chen2021wind}, DANN~\cite{ganin2016domain} and CRST~\cite{zou2019confidence} are selected as the baselines, while In+Out~\cite{chen2021wind}, MME~\cite{saito2019semi}, BiAT~\cite{jiang2020bidirectional}, and Wind~\cite{chen2021wind} are selected as the baselines for the semi-supervised domain adaptation. Out and In+Out are two straightforward ways for realizing unsupervised and semi-supervised domain adaptation, where Out means the base model is trained on the out-of-domain data (i.e., labeled source domain data) and In+Out means the base model is trained on both the in-domain and the out-of-domain data. The core of DANN is an adversarial learning algorithm that takes the domain classification loss as an auxiliary loss. CRST is also a self-training method that uses a label regularization technique to reduce the label noise from mislabeled data. WIND is a meta-learning-based domain adaptation approach that optimizes the weights of different training instances. The difference between the WIND and DaMSTF lies in that, (i) WIND only use the labeled source data to construct the meta training set, while the meta training set in the DaMSTF contains both the labeled data from the source domain and the pseudo data from the target domain. (ii) WIND does not consider the training guidance vanishment problem and the bias between the test set (i.e., target domain) and the meta validation set. \\

\subsection{Results}
\label{subsec:exp_General_Results}
To validate the effectiveness of the meta self-training, we conduct unsupervised and semi-supervised domain adaptation experiments on two benchmark datasets, i.e., BiGCN on TWITTER, and BERT on Amazon. Since the rumor detection task focuses more on the `rumor' category, we evaluate different models by their F1 score in classifying the `rumor' category. On the sentiment classification task, the prediction accuracy of different classes is equally important, so we take the macro-F1 score to evaluate different models. For semi-supervised domain adaptation, 100 labeled instances in the target domain are taken as the in-domain dataset. The experiment results are listed in Tab.~\ref{tab:TWITTERResults}, Tab.~\ref{tab:AmazonResults}.

As shown in Tab.~\ref{tab:TWITTERResults}, Tab.~\ref{tab:AmazonResults}, DaMSTF outperforms all baseline approaches on all benchmark datasets. On the rumor detection task,  DaMSTF surpasses the best baseline approaches (CRST for unsupervised domain adaptation, WIND for semi-supervised domain adaptation) by nearly 5\% on average. For the ``Fer.'' domain, where most approaches perform worse than the Out and In+Out,  DaMSTF still achieves an F1 value of 0.629, which is 40\% higher than that of the In+Out. On the sentiment classification task, DaMSTF also outperforms other approaches. Under the unsupervised domain adaptation scenario, DaMSTF surpasses the best baseline approach (DANN on the Amazon dataset) by nearly 2\% on average. Under the semi-supervised domain adaptation scenario, DaMSTF surpasses Wind, the best baseline approach on the Amazon dataset, by nearly 3\% on average. 

\subsection{Ablation Study}
\label{subsec:Ablation_Study}
This subsection presents an ablation study to understand the effectiveness of the DaMSTF. As illustrated in $\S$~\ref{sec:methodology} and $\S$~\ref{subsec:4.2}, DaMSTF combines meta-learning and self-training via two strategies: (i) expanding the meta validation set with a meta constructor; (ii) preventing the training guidance vanishment problem with a domain adversarial module. Thus, 
we separately remove the above strategies from the DaMSTF, yielding three different variants, namely DaMSTF \textit{- w/o E}, DaMSTF \textit{- w/o D}, and DaMSTF \textit{- w/o D, E}. Compared with DaMSTF, DaMSTF \textit{- w/o E} does not select examples to expand the meta validation set, which means all pseudo instances are preserved to the meta training set. DaMSTF \textit{- w/o D} removes the domain adversarial module from the DaMSTF. DaMSTF \textit{- w/o D, E} removes both two strategies. 
Other experiment settings are the same as $\S$~\ref{subsec:exp_General_Results}. 
We summarize the results in Tab.~\ref{tab:abl_TWITTER}, Tab.~\ref{tab:abl_Amazon}.

\begin{table}[bt]
\small
\centering
\begin{tabular}{c|llll|l}
\hline
                      & \multicolumn{1}{c}{Cha.} & \multicolumn{1}{c}{Fer.} & \multicolumn{1}{c}{Ott.} & \multicolumn{1}{c|}{Syd.} & Mean  \\ \hline
DaMSTF                               & 0.649                    & 0.629                    & 0.843                    & 0.731                     & 0.713 \\ \hline
\textit{- w/o D}    & 0.585                    & 0.401                    & 0.782                    & 0.724                     & 0.623 \\ \hline
\textit{- w/o E}    & 0.600                    & 0.542                    & 0.694                    & 0.685                     & 0.630 \\ \hline
\textit{- w/o D, E} & 0.569                    & 0.352                    & 0.633                    & 0.631                     & 0.547 \\ \hline
\end{tabular}
\caption{Ablation Study on TWITTER}
\label{tab:abl_TWITTER}
\end{table}

\begin{table}[bt]
\small
\centering
\setlength{\tabcolsep}{4.pt}
\begin{tabular}{c|cccc|c}
\hline
             & books & dvd   & electronics & kitchen & Mean  \\ \hline
DaMSTF                               & 0.947 & 0.935 & 0.941       & 0.947   & 0.942 \\ \hline
\textit{- w/o D}    & 0.899 & 0.917 & 0.924       & 0.935   & 0.918 \\ \hline
\textit{- w/o E}    & 0.917 & 0.929 & 0.934       & 0.945   & 0.931 \\ \hline
\textit{- w/o D, E} & 0.887 & 0.896 & 0.919       & 0.931   & 0.908 \\ \hline
\end{tabular}
\caption{Ablation Study on the Amazon dataset}
\label{tab:abl_Amazon}
\end{table}

As shown in Tab.~\ref{tab:abl_TWITTER} and Tab.~\ref{tab:abl_Amazon}, both strategies are indispensable for the effectiveness of DaMSTF, and removing either strategy can result in performance degeneration. Removing the domain adversarial learning module (DaMSTF - \textit{w/o D}) leads to an average decrease from 0.713 to 0.623 on the TWITTER dataset and from 0.942 to 0.918 on the Amazon dataset. Without expanding the meta validation set, DaMSTF - \textit{w/o E} performs worse than DaMSTF on both the TWITTER dataset (0.630 vs. 0.731 on average) and the Amazon dataset(0.931 vs. 0.942 on average). After removing both strategies, DaMSTF suffers a severe performance deterioration on both benchmark datasets.

\subsection{Effect of the unlabeled dataset size}
\begin{figure}[tbh]
  \centering
   \includegraphics[width=2.8in]{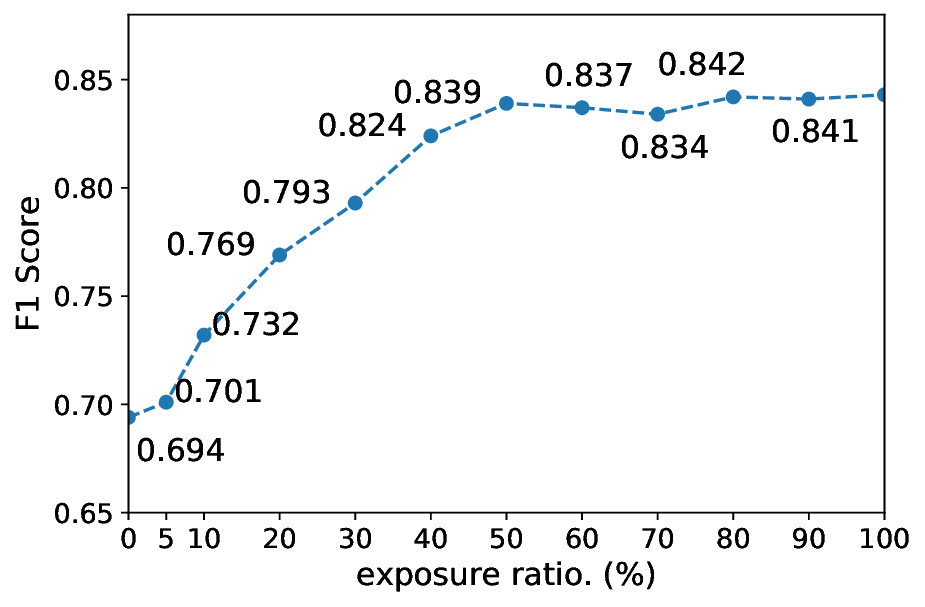}
  \caption{The impact of the size of $D^u_{T}$.}
  \label{fig:effect_of_unlabeled_size}
\end{figure}

 As illustrated in $\S$~\ref{subsec:4.2}, the second term $d_{H \Delta H}(\mathbb{D}_T, D_{T}^{l} \cup D_E)$ is close to $d_{H \Delta H}(\mathbb{D}_T, D_T^u)$ in the whole training process. From this perspective, increasing the size of the unlabeled dataset can improve the performance. To validate this, we separately expose 0\%, 5\%, 10\%, 20\%, 30\%, 40\%, 50\%, 60\%, 70\%, 80\%, 90\%, 100\% of the unlabeled data during the training. These new unlabeled dataset are denote as $D_{T}^{u}(0\%), D_{T}^{u}(5\%), \ldots, D_{T}^{u}(100\%)$ respectively. The experiments are conducted on "Ott." Domain of TWITTER and the results are presented in Fig.~\ref{fig:effect_of_unlabeled_size}.
\label{subsec:unlabeled data}

From Fig.~\ref{fig:effect_of_unlabeled_size}, we observe that the model performs poorly when using a small proportion of the unlabeled data in the training process. For example, exposing $D_T^{u}(5\%)$ to the DaMSTF only achieves an F1 score of 0.701, which is 14.2\% lower than the 0.843 achieved by exposing the $D_T^{u}(100\%)$. From 0\% to 50\%, increasing the exposure ratio consistently improves the F1 score. The improvements saturate after more than 50\% of the unlabeled data are exposed, which can be explained by the law of large numbers in the statistic theory~\cite{kraaikamp2005modern}. An exposure ratio of 50\% can be regarded as a large number for approaching the unlabeled dataset. Thus, $D_T^{u}(50\%)$ is close to $D_T^{u}(100\%)$ and $d_{H\Delta H}(\mathbb{D}_T, D_T^{u}(50\%))$ approximates $d_{H\Delta H}(\mathbb{D}_T, D_T^{u}(100\%))$,  which leads to the performance saturation.

\section{Related Work}
\label{sec:related-work}
\subsection{Domain Adaptation} 
Inspired by the taxonomy in~\citet{DBLP:conf/coling/RamponiP20}, we categorize the domain adaptation approaches into two categories: Feature-Alignment approaches and Data-Centric approaches. 
Feature-Alignment approaches~\cite{DBLP:journals/corr/TzengHZSD14, ganin2016domain,saito2019semi} focus on aligning the feature space across domains. The most well-known feature-alignment approach is DANN~\cite{ganin2016domain}, which aligns the feature space by min-max the domain classification loss. With similar efforts, MME~\cite{saito2019semi} min-max the conditional entropy on the unlabeled data. VAT~\cite{miyato2018virtual}, as well as BiAT~\cite{jiang2020bidirectional}, propose to decouple the min-max optimization process, which first imposes a gradient-based perturbation on the input space to maximize the risk loss and then minimize the final objective on the perturbed input cases.
In contrast, Data-Centric approaches exploit the unlabeled data in the target domain or select the relevant data from the source domain. 
To select relevant data, ~\cite{DBLP:conf/acl/MooreL10,DBLP:conf/acl/PlankN11} design a technique based on topic models for measuring the domain similarity.
To exploit the unlabeled data, pseudo labeling approaches, including self-training~\cite{zou2019confidence}, co-training~\cite{chen2011co}, and tri-training~\cite{saito2017asymmetric}, are widely applied and become an important direction. In the research of self-training for domain adaptation, many efforts are put into reducing the label noise of pseudo instances~\cite{zou2019confidence,zou2018unsupervised,liu2021cycle}. 
Among them, CRST~\cite{zou2019confidence} proposes a label regularization technique to reduce label noise while CST~\cite{liu2021cycle} takes Tsallis-entropy as a confidence-friendly regularize. 
In this paper, we propose to adopt meta-learning to automatically reduce label noise.
\subsection{Meta-Learning}
Meta-learning is an emerging new branch in machine learning that focuses on providing better hyperparameters for model training, including but not limited to better initial model parameters, e.g., MAML~\cite{finn2017model}, better learning rates, e.g., MetaSGD~\cite{DBLP:journals/corr/LiZCL17}, and better neural network architect, e.g., DARTs~\cite{liu2018darts}. Recent studies revealed the prospect of providing better instance weights~\cite{ren2018learning,shu2019meta,kye2020meta}. When using prototypical learning on the few-shot image classification task, MCT~\cite{kye2020meta} involves a reweighing process to obtain a more accurate class prototype.
Oriented to natural language processing tasks, ~\cite{li2020meta,chen2021wind} use the optimization-based meta-reweighting algorithm to refine the training set. Similar to DaMSTF,~\citet{wang2021meta} also proposes to combine the meta-learning algorithm and the self-training approach, but their method focuses on the neural sequence labeling task rather than the domain adaptation task. Also, they do not consider the bias between the meta-validation set and the test set, whereas reducing such bias is an important contribution of the DaMSTF. WIND~\cite{chen2021wind} is a meta-learning-based domain adaptation approach, the differences between WIND and DaMSTF are discussed in $\S$~\ref{subsec:exp_settings}. 

\section{Conclusion}
\label{sec:conclusion}
This paper proposes an improved self-training framework for domain adaptation, named DaMSTF. DaMSTF extends the basic framework for self-training approaches by involving a meta-learning module, which alleviates the label noise problem in self-training. To guarantee the effectiveness of the meta-learning module, we propose a meta constructor to improve the quality of the meta validation set, and propose a domain adversarial module to prevent the training guidance vanishment. Also, the domain adversarial learning module can align the feature space along with the self-training iterations. Extensive experiments on two popular models, BiGCN and BERT, verify the effectiveness of DaMSTF. The ablation studies demonstrate that the meta-learning module, the meta constructor, and the domain adversarial module are indispensable for the effectiveness of the DaMSTF. The limitation, ethical considerations, and social impacts of this paper are in Appendix~\ref{sec:limits} and~\ref{sec:ecsi}.

\section*{Acknowledgements} 
This work is supported by the following foundations: the National Natural Science Foundation of China under Grant No. 62025208, the Xiangjiang Laboratory Foundation under Grant No. 22XJ01012, 2022 International Postdoctoral Exchange Fellowship Program (Talent-Introduction Program) under Grant No. YJ20220260.
\bibliography{anthology}

\appendix
\newpage
\section{Proof For Theorem~\ref{theo:1}}
\label{sec:appendix_A}

\setcounter{theorem}{0}
\begin{theorem}
\label{theo:1}
Let $\mathbf{w}_{i}$ be the weight of the training instance $i$, denoted as $(x_i, y_i)$, in $\mathcal{B}$, the gradient of $\mathbf{w}_{i}$ on $\mathcal{L}_M$ can be represented by the similarity between the gradients on training instance $i$ and the gradients on the meta validation set: 

\begin{small}
\begin{equation}
\frac{\partial L_M(\hat{\mathbf{\theta}}(\mathbf{w}))}{\partial \mathbf{w}_i} = -\frac{\eta}{|\mathcal{B}|}\cdot [\frac{1}{|D_M|}\sum_{j=1}^{|D_M|} \vec{\mathbf{g}}_{\hat{\mathbf{\theta}}}(x_j, y_j)^T] \cdot \vec{\mathbf{g}}_{\mathbf{\theta}}(x_i, y_i) \nonumber
\end{equation}
\end{small}
\noindent where $\frac{1}{|D_M|}\sum_{j=1}^{|D_M|} \vec{\mathbf{g}}_{\hat{\mathbf{\theta}}}(x_j, y_j)^T$ is the gradients of $\hat{\mathbf{\theta}}$ on $D_M$, $\vec{\mathbf{g}}_{\mathbf{\theta}}^i(x_i, y_i)$ is the gradients of $\mathbf{\theta}$ on the training instance $i$, $\eta$ is the learning rate in Eq.~\eqref{eq:theta_hat}
\end{theorem}

\begin{proof}
Based on Eq.~\eqref{eq:inner} and Eq.~\eqref{eq:theta_hat} in $\S$~\ref{subsec:meta_learning_module}, we conclude the pseudo updated parameters $\hat{\mathbf{\theta}}(\mathbf{w})$ as:

\begin{small}
\begin{equation}
\hat{\mathbf{\theta}}(\mathbf{w}) = \mathbf{\theta} - \eta \cdot \frac{1}{|\mathcal{B}|} \cdot\sum_{x_i, y_i \in \mathcal{B}} \sigma{(\mathbf{w}_i)} \cdot \frac{\partial \mathcal{E}(\Phi(x_i;\mathbf{\theta}), y_i)}{\partial \mathbf{\theta}} \label{eq:new_hat_theta}
\end{equation}
\end{small}

\noindent We then take the gradient of $\mathbf{w}_i$ on $\hat{\mathbf{\theta}}(\mathbf{w})$ as:

\begin{small}
\begin{equation}
\frac{\partial \hat{\mathbf{\theta}}(\mathbf{w})}{\partial \sigma(\mathbf{w}_i)} = -\frac{\eta}{|\mathcal{B}|} \cdot \frac{\partial \mathcal{E}(\Phi(x_i;\mathbf{\theta}), y_i)}{\partial \mathbf{\theta}} \label{eq:wi_grad}
\end{equation}
\end{small}

\noindent Based on Eq.~\eqref{eq:wi_grad}, we derivate the gradient of $\mathbf{w}_i$ on $\mathcal{L}_M$ as:
\begin{small}
\begin{eqnarray}
\frac{\partial L_M(\hat{\mathbf{\theta}}(\mathbf{w}))}{\partial \mathbf{w}_i} &=& [\frac{\partial L_M(\hat{\mathbf{\theta}}(\mathbf{w}))}{\partial \hat{\mathbf{\theta}}(\mathbf{w})}]^T \cdot [\frac{\partial \hat{\mathbf{\theta}}(\mathbf{w})}{\partial \sigma(\mathbf{w}_i)}]\cdot [\frac{\partial \sigma(\mathbf{w})}{\partial \mathbf{w}}] \nonumber\\
& = & [\frac{1}{|D_M|} \cdot \sum_{j=1}^{|D_M|}\frac{\partial \mathcal{E}(\Phi(x_j;\hat{\mathbf{\theta}}(\mathbf{w})),y_j)}{\partial \hat{\mathbf{\theta}}(\mathbf{w})}]^T \cdot \nonumber \\
 && \quad[- \frac{\eta}{|\mathcal{B}|} \cdot \frac{\partial \mathcal{E}(\Phi(x_i;\mathbf{\theta}), y_i)}{\partial \mathbf{\theta}}]\cdot\nonumber\\
 &&\quad [\sigma(\mathbf{w}_i)(1-\sigma(\mathbf{w}_i))] \nonumber\\
& = & -\frac{\eta\sigma(\mathbf{w}_i)(1-\sigma(\mathbf{w}_i))}{|\mathcal{B}|}\cdot \nonumber \\
&& [\frac{1}{|D_M|}\sum_{j=1}^{|D_M|} \vec{\mathbf{g}}_{\hat{\mathbf{\theta}}}(x_j, y_j)^T] \cdot \nonumber\\ 
&& \vec{\mathbf{g}}_{\mathbf{\theta}}(x_i, y_i)  \label{eq:partial_wi}
\end{eqnarray}
\end{small}

\noindent where the second line is obtained by substituting $\mathcal{L}_M$ and $\hat{\mathbf{\theta}}$ with Eq.~\eqref{eq:L_meta} and Eq.~\eqref{eq:new_hat_theta}. Substitute $\vec{\mathbf{g}}_{\hat{\mathbf{\theta}}}(x_j, y_j) = \frac{\partial \mathcal{E}(\Phi(x_j;\hat{\mathbf{\theta}}(\mathbf{w})),y_j)}{\partial \hat{\mathbf{\theta}}(\mathbf{w})}$ and $\vec{\mathbf{g}}_{\mathbf{\theta}}(x_i, y_i) = \frac{\partial \mathcal{E}(\Phi(x_i;\mathbf{\theta}), y_i)}{\partial \mathbf{\theta}}$ and rearrange the terms, we obtain the third line.
\noindent The proof of Theorem~\ref{theo:1} is completed.
\end{proof}

\section{Proof For Theorem~\ref{theo:2} and Theorem~\ref{theo:3}}
\label{sec:appendix_B}

\begin{definition}
\label{Def:1}
\textit{disagreement} is a measure to quantify the different performances of two different hypotheses on a specific dataset. Denote the two hypotheses as $h_1$ and $h_2$, and denote the specific dataset as $D$, then the {disagreement} of $h_{1}$ and $h_{2}$ on $D$ is formulated as:
\begin{small}
\begin{equation}
\epsilon_D(h_1, h_2) = \frac{1}{|D|} \sum_{i=1}^{|D|}[ \frac{1}{C}*|| h_1(x) - h_2(x) ||_1] \label{eq:disagreement}
\end{equation}
\end{small}
\noindent where $C$ is the number of classes, $h_1(x)$ and $h_2(x)$ are one-hot vectors representing the models' predictions.
\end{definition}

\begin{definition}
\label{Def:2}
$H{\Delta}H$-distance is a metric for evaluating the divergence of the data distribution between two datasets. Formally, $H{\Delta}H$-distance is computed as:
\begin{small}
\begin{equation}
d_{\mathcal{H}\Delta\mathcal{H}}(D_1, D_2) = 2 \mathop{sup}\limits_{h_1, h_2 \in \mathcal{H}} | \epsilon_{D_1}(h1, h2) - \epsilon_{D_2}(h1, h2)| \label{eq:HH-distance}
\end{equation}
\end{small}
\noindent where $H$ is the hypothesis space and $sup$ denotes the supremum. 
\end{definition}

The concepts \textit{disagreement} and $H{\Delta}H$-distance are introduced in Definition~\ref{Def:1} and Definition~\ref{Def:2}, respectively.
Based on the \textit{disagreement} and $H{\Delta}H$-distance, the proof for Theorem~\ref{theo:2} is presented as below.
\setcounter{theorem}{0}
\begin{lemma}
\label{lemma:1}
Assume there exists two dataset, i.e., $D_1$, $D_2$. Let $X_1 = \lbrace x_i | (x_i, y_i) \in D_1 \rbrace$ and $X_2 = \lbrace x_i | (x_i, y_i) \in D_2 \rbrace$ denotes the set of input case from $D_1$ and $D_2$. If  $X_1 \subseteq X_2$, then
$$d_{H\Delta H}(D_1, D_2) = 2\cdot \frac{|D_2| - |D_1|}{|D_2|}$$
holds.
\end{lemma}
\begin{proof}
Let $\mathit{I}_k(h_1, h_2) = \frac{1}{C}*||h_1(x_k) - h_2(x_k)||_1$ denote the difference of two hypothesis $h_1$ and $h_2$ on instance $x_k$, then the \textit{disagreement} of $h_1$ and $h_2$ on the dataset $D$ can be rewritten as: \\
$$
\epsilon_D(h_1, h_2) = \frac{1}{|D|} \sum_{i=1}^{|D|} \mathit{I}_i(h_1, h_2)
$$
Based on the Definition~\ref{Def:2}, the $H\Delta H$ distance between $D_1$ and $D_2$ is as below:
\begin{small}
\begin{equation}
d_{H\Delta H}(D_1, D_2) = 2 \mathop{sup}\limits_{h_1, h_2 \in \mathcal{H}} | \epsilon_{D_1}(h1, h2) - \epsilon_{D_1}(h1, h2)|
\end{equation}
\end{small}

Expanding the item $\epsilon_{D_1}(h1, h2)$ and $\epsilon_{D_1}(h1, h2)$, we can obtain:

\begin{small}
\begin{eqnarray}
&&|\epsilon_{D_2}(h1, h2) - \epsilon_{D_1}(h1, h2)|  \nonumber \\
&& =|\frac{1}{|X_2|}\sum_{x_i\in X_2}\mathit{I}_i(h_1, h_2) - \frac{1}{|X_1|}\sum_{x_i\in X_1}\mathit{I}_i(h_1, h_2)| \nonumber \\
&& = |\frac{|X_1|}{|X_2|}*\frac{1}{|X_1|}\sum_{x_i \in X_1}\mathit{I}_i(h_1, h_2) \nonumber \\
&&\quad+ \frac{|\bar{X}_1|}{|X_2|}*\frac{1}{|\bar{X}_1|}\sum_{x_k \in \bar{X}_1}\mathit{I}_i(h_1, h_2) \nonumber \\
 &&\quad - \frac{1}{|X_1|}\sum_{x_i \in X_1}\mathit{I}_i(h_1, h_2)| \nonumber \\
&& = |\frac{1}{|X_2|}\sum_{x_k\in\bar{X}_1} \mathit{I}_k(h_1, h_2) \nonumber \\
&&\quad- \frac{|X_2| - |X_1|}{|X_2|}\cdot\frac{1}{|X_1|}\sum_{x_i\in X_1} \mathit{I}_i(h_1, h_2) | \nonumber \\
&& = \frac{1}{|X_2|}|\sum_{x_k \in \bar{X}_1} \mathit{I}_k(h_1, h_2) \nonumber\\
&&\quad- \frac{|\bar{X}_1|}{|X_1|}\cdot\sum_{x_i \in X_1} \mathit{I}_i(h_1, h_2)| \nonumber \\
&& = \frac{|\bar{X}_1|}{|X_2|} |\epsilon_{\bar{D}_1}(h_1, h_2) - \epsilon_{D_1}(h_1, h_2)| \label{eq:dHH_item}
\end{eqnarray}
\end{small}
where $\bar{X}_1$ is the complement set of $X_1$ in $X_2$, i.e, $\bar{X}_1 = X_2 - X_1$. Correspondingly, $\bar{D}_1 = \lbrace (x_i, y_i) | (x_i, y_i) \in D_2 \textnormal{ and } x_i \in \bar{X} \rbrace$, and thus $|\bar{X}_1| = |\bar{D}_1|$ holds.

As $0 \leq \epsilon_{\bar{D}_1}(h_1, h_2) \leq 1$ and $0 \leq \epsilon_{D_1}(h_1, h_2) \leq 1$ , we conclude the inequation below:
\begin{equation}
|\epsilon_{\bar{D}_1}(h_1, h_2) - \epsilon_{D_1}(h_1, h_2)| \leq 1
\label{eq:divergence}
\end{equation}
 
Since $D_1$ and $\bar{D}_{1}$ do not overlap, $\epsilon_{\bar{D}_1}(h_1, h_2)$ is independent of $\epsilon_{D_1}(h_1, h_2)$. Thus, we can maximize the left term in inequation~\eqref{eq:divergence} by finding two hypotheses $\hat{h}_{1}$ and $\hat{h}_{2}$, which make $\epsilon_{\bar{D}_1}(\hat{h}_1, \hat{h}_2) = 1$ and $\epsilon_{D_1}(\hat{h}_1, \hat{h}_2)=0$. Thus,

\begin{small}
\begin{eqnarray}
&&d_{H\Delta H}(D_1, D_2) \nonumber \\
&=& 2 \mathop{sup}\limits_{h_1, h_2 \in \mathcal{H}} | \epsilon_{D_2}(h1, h2) - \epsilon_{D_1}(h1, h2)| \nonumber\\
&=& 2\cdot \frac{|\bar{X}_1|}{|X_2|} \mathop{sup}\limits_{h_1, h_2 \in \mathcal{H}}|\epsilon_{\bar{D}_1}(h_1, h_2) - \epsilon_{D_1}(h_1, h_2)|\nonumber \\
&=& 2\cdot \frac{|\bar{D}_1|}{|D_2|} \mathop{sup}\limits_{h_1, h_2 \in \mathcal{H}}|\epsilon_{\bar{D}_1}(h_1, h_2) - \epsilon_{D_1}(h_1, h_2)|\nonumber \\
&=& 2\cdot \frac{|\bar{D}_1|}{|D_2|} |\epsilon_{\bar{D}_1}(\hat{h}_1, \hat{h}_2) - \epsilon_{D_1}(\hat{h}_1, \hat{h}_2)|\nonumber \\
&=& 2\cdot \frac{|\bar{D}_1|}{|D_2|} \nonumber \\
&=& 2\cdot \frac{|D_2| - |D_1|}{|D_2|} \nonumber
\end{eqnarray}
\end{small}
The proof of Lemma 1 is completed. 
\end{proof}

\setcounter{theorem}{1}
\begin{theorem}
Assume there exists an ideal hypothesis, denoted as $h^*$, which correctly map all instances in the target domain to their groud-truth labels. In the self-training iteration $t$, let $\epsilon_{D_{T}^{l}}(h^t)$ and $\epsilon_{D_E}(h^t)$ be the error rate of the hypothesis $h^{t}$ on $D^l_{T}$ and $D_E$, respectively. Then, the error rate of the hypothesis $h^{t}$ on the target domain is upper bounded by: 

\begin{small}
\begin{eqnarray}
\epsilon_{\mathbb{D}_T}(h^t) &\leq& \epsilon_{D_{T}^{l} \cup D_E}(h^t) + \frac{1}{2}d_{H \Delta H}(\mathbb{D}_T, D_{T}^{l} \cup D_E) \nonumber \\
&& \quad + \rho \cdot \epsilon_{D_E}(h^*, h^{t-1}) 
\end{eqnarray}
\end{small}

\noindent where $\rho = \frac{|D_E|}{|D_{T}^{l}|+|D_E|}$ is a coefficient related to the size of $D^l_{T}$ and $D_{E}$, $\epsilon_{D_{T}^{l} \cup D_E}(h^t)$ is the error rate of the hypothesis $h^{t}$ on the union of $D_{T}^{l}$ and $D_E$.
\end{theorem}
\begin{proof}
In the meta-learning module, the final objective is to minimize the risk loss on the meta validation set $D_{T}^{l} \cup D_E$. Thus, according to the learning theory~\cite{ben2010theory}, the upper bound of the error rate on the test set (i.e., the target domain) is:

\begin{small}
\begin{eqnarray}
\epsilon_{\mathbb{D}_T}(h^t) &\leq& \epsilon_{D_{T}^{l} \cup D_E}(h^t) + \frac{1}{2}d_{H \Delta H}(\mathbb{D}_T, D_{T}^{l} \cup D_E) \nonumber \\
&& \quad+ \epsilon_{\mathbb{D}_T}(h^*) + \epsilon_{D_{T}^{l} \cup D_E}(h^*)
\label{eq:init_ineq}
\end{eqnarray}
\end{small}

\noindent Because $h^{*}$ is an ideal hypothesis on the target domain, $\epsilon_{\mathbb{D}_T}(h^*) = 0$ holds true. 

Expanding $\epsilon_{D_{T}^{l} \cup D_E}(h^*)$ with the definition in Eq.~\eqref{eq:disagreement},
\begin{small}
\begin{eqnarray}
&&\epsilon_{D_{T}^{l} \cup D_E}(h^*) \nonumber\\
&=& \frac{1}{|D_{T}^{l}|+|D_E|} \sum_{(x, y) \in D_{T}^{l} \cup D_E}[ \frac{1}{C}*|| h^*(x) - y ||_1] \nonumber \\
&=& \frac{1}{|D_{T}^{l}|+|D_E|} \lbrace \sum_{(x, y) \in D_{T}^{l}}[ \frac{1}{C}*|| h^*(x) - y ||_1] \nonumber\\
&& \quad \quad \quad \quad \quad+ \sum_{(x, y) \in D_E}[ \frac{1}{C}*|| h^*(x) - y ||_1]\rbrace \nonumber \\
&=& \frac{1}{|D_{T}^{l}|+|D_E|} \lbrace |D_{T}^{l}|\cdot\epsilon_{D_{T}^{l}}(h^*) + |D_E|\cdot \epsilon_{D_E}(h^*)\rbrace \nonumber\\
\label{eq:union_error_rate}
\end{eqnarray}
\end{small}

\noindent Substituting Eq.~\eqref{eq:union_error_rate} into Eq.~\eqref{eq:init_ineq}, we have:

\begin{small}
\begin{eqnarray}
&&\epsilon_{\mathbb{D}_T}(h^t) \nonumber \\
&\leq& \epsilon_{D_{T}^{l} \cup D_E}(h^t) + \frac{1}{2}d_{H \Delta H}(\mathbb{D}_T, D_{T}^{l} \cup D_E) + \epsilon_{\mathbb{D}_T}(h^*) \nonumber \\
&&+ \frac{1}{|D_{T}^{l}|+|D_E|} \lbrace |D_{T}^{l}|\cdot\epsilon_{D_{T}^{l}}(h^*) + |D_E|\cdot \epsilon_{D_E}(h^*)\rbrace \nonumber \\
\label{eq:proof}
\end{eqnarray}
\end{small}

\noindent For any instance $(x, y) \in D_E$, $y$ is the pseudo label, i.e., the prediction of hypothesis $h^{t-1}$. Thus, we have:

\begin{small}
\begin{eqnarray}
&& \epsilon_{D_E}(h^*) \nonumber \\
&=& \frac{1}{|D_E|}\sum_{(x, y) \in D_E}[ \frac{1}{C}*|| h^*(x) - y ||_1] \nonumber \\
&=&\frac{1}{|D_E|}\sum_{(x, y) \in D_E}[ \frac{1}{C}*|| h^*(x) - h^{t-1}(x) ||_1] \nonumber\\
&=&\epsilon_{D_E}(h^*, h^{t-1}) 
\end{eqnarray}
\end{small}

Since $D_{T}^{l}$ is a subset of $\mathbb{D}_T$, $\epsilon_{D_{T}^{l}}(h^*) = 0$ holds true.
By eliminating $\epsilon_{\mathbb{D}_T}(h^*)$ and $\epsilon_{D_{T}^{l}}(h^*)$ in Eq.\eqref{eq:proof}, and substituting $\epsilon_{D_E}(h^*)$ with $\epsilon_{D_E}(h^*, h^{t-1})$, we have:

\begin{small}
\begin{eqnarray}
\epsilon_{\mathbb{D}_T}(h^t) &\leq& \epsilon_{D_{T}^{l} \cup D_E}(h^t) + \frac{1}{2}d_{H \Delta H}(\mathbb{D}_T, D_{T}^{l} \cup D_E) \nonumber \\
&& \quad \quad + \frac{|D_E|}{|D_{T}^{l}|+|D_E|} \cdot \epsilon_{D_E}(h^*, h^{t-1})\rbrace \nonumber
\end{eqnarray}
\end{small}

The proof of Theorem 2 is completed.
\end{proof}

\begin{theorem}
\label{theo:3}
Assume there exists three datasets, $D_{1}$, $D_{2}$, $D_{3}$,  and let $X_1$, $X_2$, $X_3$ denotes the set of input cases in these three datasets, i.e., $X_1 = \lbrace x_i|(x_i, y_i) \in D_1 \rbrace$, $X_2 = \lbrace x_i|(x_i, y_i) \in D_2 \rbrace$, $X_3 = \lbrace x_i|(x_i, y_i) \in D_3 \rbrace$. If $X_{1}\subseteq X_{2} \subseteq X_{3}$, then 
$$d_{H\Delta H}(D_2, D_3) \leq d_{H\Delta H}(D_1, D_3)$$
holds
\end{theorem}
\begin{proof}
According to Lemma~\ref{lemma:1}, 
$$d_{H\Delta H}(D_2, D_3)=2\cdot\frac{|D_3| - |D_2|}{|D_3|}$$
$$d_{H\Delta H}(D_1, D_3)=2\cdot\frac{|D_3| - |D_1|}{|D_3|}$$
Since $X_{1}\subseteq X_{2}$, $|D_1| \leq |D_2|$ holds. Thus, 
$$d_{H\Delta H}(D_2, D_3) < d_{H\Delta H}(D_1, D_3)$$
holds.\\
The proof of Theorem~\ref{theo:3} is completed. 
\end{proof}

\section{Implementation Details}
\label{sec:appendix_C}
The base model on the rumor detection task is BiGCN~\cite{bian2020rumor}, while the base model on the sentiment classification task is BERT~\cite{devlin2019bert}. On the benchmark datasets, we conduct domain adaptation experiments on every domain. When one domain is taken as the target domain for evaluation, the rest domains are merged as the source domain. 
For example, when the ``books” domain in the Amazon dataset is taken as the target domain, the ``dvd”, ``electronics” and ``kitchen” domains are merged as the source domain.

The unlabeled data from the target domain are used for training the model, and the labeled data from the target domain are used for testing and validating the model (with a ratio of 7:3). Notes that the TWITTER dataset does not contain extra unlabeled data, we take 70\% of the labeled data on the target domain as the unlabeled data for training, and the rest will be preserved for testing and validating. The experiments on TWITTER are conducted on ``Cha.”, ``Fer.”, ``Ott.”, and ``Syd.”\footnote{The labeled data in ``Ger.” domain is too scare to provide extra unlabeled data.}. 

The implementation of BiGCN to realize the rumor detection task is provided in~\cite{bian2020rumor}, and we follow the description in~\cite{bian2020rumor} to train the BiGCN model with the TWITTER dataset. The implementation of BERT to realize the sentiment analysis task can be found in~\cite{devlin2019bert}. We download the pre-trained BERT from \url{https://huggingface.co/bert-base-uncased}\footnote{under the license apache-2.0} and fit the BERT on the Amazon dataset with the instruction in~\cite{devlin2019bert}. Since DANN, FixMatch, CST, MME, WIND, and BiAT are model agnostic, we implement them according to the cited references~\cite{ganin2016domain,sohn2020fixmatch,liu2021cycle,saito2019semi,chen2021wind,wang2019bilateral}.
For the symbols in Algorithm~\ref{algo:meta self-training}, we set $\mathcal{T}_{M}$ as 5, $\mathcal{T}_{D}$ as 5, $\mathcal{T}_G$ as 1. We set $\eta_1$ and $\eta_2$ in Algorithm~\ref{algo:meta self-training} as $5e-4$ and $5e-3$ for the BiGCN model, and as $5e-6$ and $2e-5$ for the BERT model. We set $\eta$ in Eq.~\eqref{eq:theta_hat} as $5e-5$ for the BERT model, and $5e-3$ for the BiGCN model. We set $\gamma$ in Eq.~\eqref{eq:w_update_A} as 0.1 for both the BERT and the BiGCN model. We conduct all experiments the GeForce RTX 3090 GPU with 24GB memory. 

\begin{figure*}[t]
  \centering
   \includegraphics[width=5.2in]{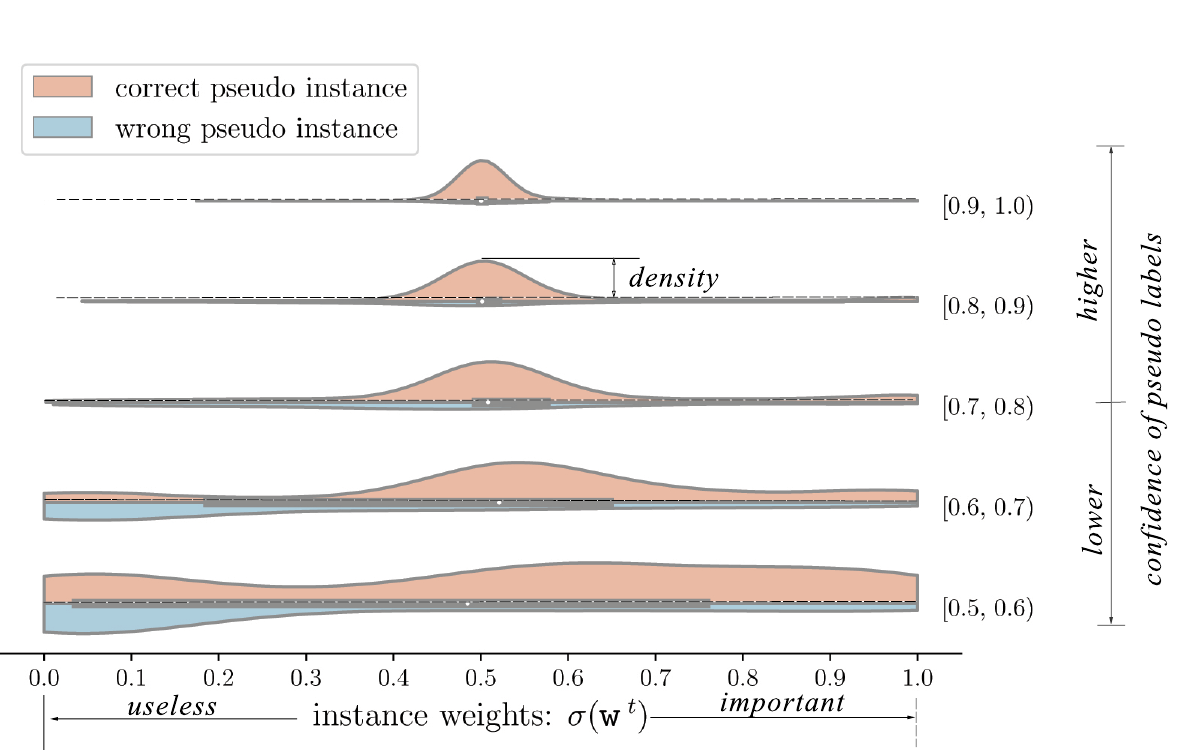}
  \caption{Distribution of activated weights ($\sigma(\mathbf{w}^t))$) over different kinds of pseudo instances.}
  \label{fig:reweighting}
\end{figure*}

\section{Statistics of the Datasets}
\label{sec:dataset}
TWITTER dataset is provided in the \href{https://figshare.com/ndownloader/articles/6392078/versions/1}{site}\footnote{https://figshare.com/ndownloader/articles/6392078/} under a CC-BY license. Amazon dataset is accessed from \href{https://github.com/ruidan/DAS}{https://github.com/ruidan/DAS}. The statistics of the TWITTER dataset and the Amazon dataset is listed in Table~\ref{tab:statisticTWITTER} and Table~\ref{tab:Amazon_sts}.
\begin{table}[t]
\small
\centering
\setlength{\tabcolsep}{2.pt}
\begin{tabular}{l|lll}
\hline
Domain            & Rumours        & Non-Rumours    & Total \\ \hline
Charlie Hebdo\#   & 458 (22\%)     & 1,621 (78\%)   & 2,079 \\
Ferguson\#        & 284 (24.8\%)   & 859 (75.2\%)   & 1,143 \\
Germanwings Crash & 238 (50.7\%)   & 231 (49.3\%)   & 469   \\
Ottawa Shooting   & 470 (52.8\%)   & 420 (47.2\%)   & 890   \\
Sydney Siege      & 522 (42.8\%)   & 699 (57.2\%)   & 1,221 \\ \hline
Total             & 1,921 (34.0\%) & 3,830 (66.0\%) & 5,802 \\ \hline
\end{tabular}
\caption{Statistics of the TWITTER dataset.}
\label{tab:statisticTWITTER}
\end{table}

\begin{table}[t]
\small
\centering
\begin{tabular}{l|ll|l}
\hline
Domains     & positive    & negative    & unlabeled \\ \hline
books       & 1000 (50\%) & 1000(50\%)  & 6001      \\
dvd         & 1000 (50\%) & 1000 (50\%) & 34,742    \\
electronics & 1000 (50\%) & 1000 (50\%) & 13,154    \\
kitchen     & 1000 (50\%) & 1000 (50\%) & 16,786    \\ \hline
\end{tabular}
\caption{Statistics of the Amazon dataset}
\label{tab:Amazon_sts}
\end{table}

\section{Extra Experiments}
\label{sec:extra_exp}
\subsection{Instance Reweighting}
\label{subsec:meta_reweight_exp}
To investigate the effectiveness of the meta-learning module, we conduct an experiment to visualize the optimized instance weights on different pseudo instances. In detail, the experiments are conducted on the 'Cha.' domain of the TWITTER dataset. Since the unlabeled data in the TWITTER dataset is constructed with the labeled data in the target domain (illustrated in $\S$~\ref{sec:experiments}),  we are aware of the pseudo labels' correctness. Thus, we can visualize the relevance among the instance weights, pseudo labels' correctness, and pseudo labels' confidence, the experiment results are shown in Fig.~\ref{fig:reweighting}. 

Fig.~\ref{fig:reweighting} is a violin plot in a horizontal direction, where each curve represents a distribution of the instance weights. The height of the curve represents the probability density. In each confidence interval, the yellow curve is the distribution over the correct pseudo instances while the blue curve is the distribution over the wrong pseudo instances. It should be noted that the probability density is normalized in each confidence interval. Thus, the area of the two kinds curves is equal to 1.0 in each confidence interval. From Fig.~\ref{fig:reweighting}, we can obtain the following observations.

Firstly, the meta-learning module is effective in reducing label noise. In different confidence intervals, especially in [0.5-0.6] and [0.6-0.7], the peak of the blue curve is smaller than 0.2, meaning that the wrong pseudo instances are mainly allocated low instance weights. Thus, the adverse impact from the wrong pseudo instances is reduced.

Secondly, larger instance weights are allocated to the correct pseudo instances with low confidence. In specific, large instance weights (i.e., >0.5) mainly appears in the bottom two sub-graph,  so the large instance weights are mainly allocated to the correct pseudo instances whose confidence is lower than 0.7. Thus,  the meta-learning module is also effective in mining hard pseudo examples.

\subsection{Error rates on the expansion examples} 
\label{subsubsec:err_exp}
\begin{figure}[b]
  \centering
   \includegraphics[height=1.5in]{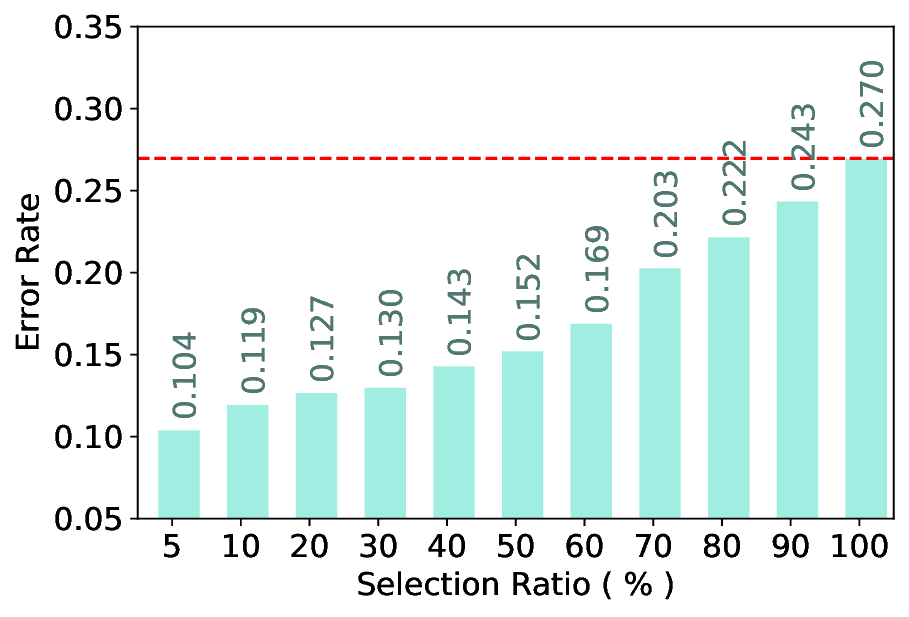}
  \caption{Error rate on the examples with different prediction entropy.}
  \label{fig:error_entropy}
\end{figure}

According to Theorem~\ref{theo:2} in $\S$~\ref{sec:theoretical_analysis}, the performance of the DaMSTF is limited by the error rate of the expansion examples, i.e., $\epsilon_{D_E}(h^*, h^{t-1})$. By selecting the examples with the lowest prediction entropy as the expansion example, the meta constructor can reduce $\epsilon_{D_E}(h^*, h^{t-1})$, thereby can improve the performance of the DaMSTF. In this subsection, we examine the reliability of the meta constructor, i.e., visualizing the relationship between the prediction entropy and the prediction correctness. Specifically, we first compute and sort the prediction entropy on the ``Syd.'' domain. We then select the top 5\%, 10\%, 20\%, 30\%, 40\%, 50\%, 60\%, 70\%, 80\%, 90\%, 100\% of the pseudo instances to compute the error rate between the selected predictions and their ground-truth labels. We summarize the experiment results in Fig.~\ref{fig:error_entropy}.

\subsection{Risk loss on the expansion examples} 
\label{subsec:loss_on_expan}
\begin{figure}[tbh]
  \centering
   \includegraphics[height=1.5in]{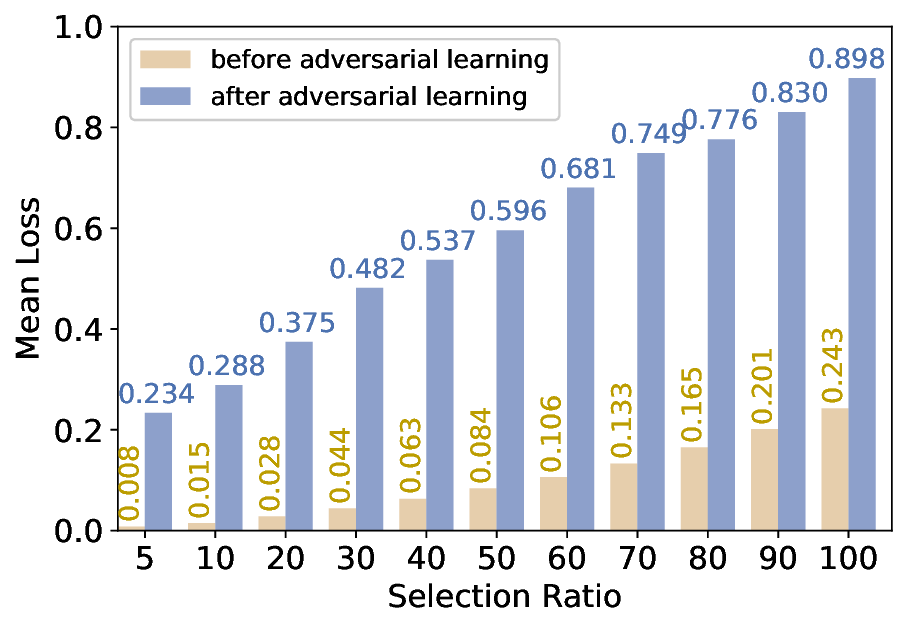}
  \caption{Risk loss on the examples with different prediction entropy.}
  \label{fig:ExpandLoss}
\end{figure}
As discussed in $\S$~\ref{subsec:val_exp}, expanding the meta validation set is challenged by the training guidance vanishment problem, since the model's risk loss, as well as the model's gradient, on the expansion examples is negligible. As a complementary, we design a domain adversarial learning module to perturb the model's parameters, thereby increasing the model's gradients on the expansion examples. Here, we provide an intuitive explanation for the necessity of introducing domain adversarial learning. Specifically, we exhibit the relationship between the predictive entropy and the risk loss, and present the changes of the risk loss before and after the parameters perturbation. The experimental settings are the same as $\S$~\ref{subsubsec:err_exp}, and we summarize the results in Fig.~\ref{fig:ExpandLoss}.

From Fig.~\ref{fig:ExpandLoss}, we observe that the mean risk loss decreases along with the decrease of the selection rate, and the risk loss on the examples with small predictive entropy is negligible. On the examples with the lowest 10\% predictive entropy (i.e., expansion examples in our setting), the mean risk loss is only 0.015. Considering that the gradient is back-propagated from the risk loss, these expansion examples cannot produce acceptable gradients. Accordingly, these expansion examples cannot provide indicative training guidance. 
After perturbing the model parameters with the domain adversarial learning module, the risk loss on the expansion examples (Selection Ratio=0.1) sharply increases from 0.015 to 0.288. Thus, the domain adversarial learning module is an indispensable complement to the meta constructor.

\section{Limitation}
\label{sec:limits}
Although our approach produces promising results on two datasets, there are certain limitations. In the future, we will continue to dig into these concerns.

Firstly, we evaluate the DaMSTF on two classification tasks. We do not conduct experiments on other NLP tasks, such as machine translation~\cite{yang2018unsupervised} or named entity recognition~\cite{jia2019cross}. Nonetheless, as text classification is a fundamental task, other NLP applications can be specified as a case of classification. For example, named entity recognition can be formulated as a word-word relation classification task~\cite{li2022unified}. 
  

Secondly, the meta-learning module carries out extra computation overhead. As the bi-level hyper-parameters optimization involves a second-order derivate on the model’s parameters, their computation overhead is quadratic to the model’s parameters. In DaMSTF, we use the approximation techniques in WIND to compute the derivate, which is linear to the model’s parameters. In the future, we will investigate other techniques to accelerate the DaMSTF.

\section{Ethical considerations and social impacts}
\label{sec:ecsi}
This paper involves the use of existing artifact(s), including two benchmark datasets and the pre-trained BERT model. Their intention for providing the artifacts is to inspire the following research, our use is consistent with their intended use.

Rumor, as well as rumor detection, is very sensitive for the social order. In this paper, we conduct experiments on a rumor detection task and prepare to release the code in the future. Since the model's prediction is not that reliable, it may lead to social harm when the model's error prediction is used with malicious intentions. For example, people may use the model's error prediction as support evidence, so as to deny a correct claim or to approve a rumor claim. Here, we seriously declare that the model's prediction cannot be taken as the support evidence. In the released code, we will constrain the input format of the model, making unprofessional individuals unable to directly use the model.

\end{document}